\documentclass{IOS-Book-Article}

\usepackage{mathptmx}
\usepackage{times} % Kept for legacy support, though mathptmx generally supersedes it

\usepackage{amsmath, amssymb, amsthm}

\usepackage{graphicx}
\usepackage{pgfplots}
\usepackage{tikz}
\usetikzlibrary{arrows.meta, positioning, shapes, quotes}
\pgfplotsset{compat=1.18}

\usepackage{booktabs}
\usepackage{tabularx}
\usepackage{array}
\usepackage{multirow}

\usepackage{caption}
\usepackage{subcaption} % Used instead of subfig
\usepackage{float}
\usepackage{stfloats}

\usepackage{xcolor} % Explicitly loaded for text colors
\usepackage{soul}
\setuldepth{article}
\usepackage{listings}
\usepackage{url}
\usepackage{hyperref}
\usepackage{algorithm}
\usepackage{algorithmic}
\usepackage[switch]{lineno}
\usepackage{ragged2e}

\newtheorem{example}{Example}

\newtheorem{definition}{Definition}

\newcommand{\embed}{{\sf Emb}}
\newcommand{\nli}{{\sf NLinf}}
\newcommand{\similar}{{\sf Sm}}
\newcommand{\prompt}{{\sf Inst}}
\newcommand{\myabstract}{{\sf Tran}}

\newcommand{\BibTeX}{B\kern-.05em\textsc{i\kern-.025em b}\kern-.08em\TeX}

\def\hb{\hbox to 11.5 cm{}}

\begin{document}
\pagestyle{headings}
\def\thepage{}
\begin{frontmatter}              % The preamble begins here.

\title{Identifying Implicit Premises for Logical Reconstruction of Argument Graphs}

\markboth{}{April 2026\hb}
%\subtitle{Subtitle}

\author[A]{\fnms{Xuyao} \snm{Feng}\orcid{0009-0009-0888-6233}\thanks{Corresponding Author: Xuyao Feng, xuyao.feng.20@ucl.ac.uk.}}
and
\author[A]{\fnms{Anthony} \snm{Hunter}\orcid{0000-0001-5602-7446}}
\runningauthor{X. Feng et al.}
\address[A]{Department of Computer Science, University College London, United Kingdom}
\begin{abstract}
The logical reconstruction of argument graphs from natural language text is challenging because of the prevalence of enthymemes (i.e., arguments with implicit premises). There are natural language processing methods for identifying enthymemes in text, and there are symbolic methods based on abduction for identifying missing premises in a logical representation of enthymemes. However, there is a need for methods to generate implicit premises to logically show a known entailment or contradiction relationship between a pair of statements. To address this, we propose a neuro-symbolic pipeline that uses large language models (LLMs) to generate intermediate implicit premises that are translated into logical formulae and used with logical formulae representing explicit premises and explicit claims to show the logical relationships between them (entailment, contradiction, or neutrality). 
Our approach is evaluated on the Microtext Argumentative Corpus.
\end{abstract}
\end{frontmatter}
\markboth{April 2026\hb}{April 2026\hb}

\section{Introduction}
\label{section:introduction}

The logical reconstruction of argument graphs from natural language text is a fundamental challenge in computational argumentation. Such reconstructions aim to formalize the underlying inferential structure of arguments, representing components (premises, claims) as nodes and their semantic relationships—such as \emph{entailment} (support), \emph{contradiction} (attack), or \emph{neutrality}—as edges in a graph. However, this task is complicated by the widespread occurrence of \textbf{enthymemes}: arguments that are logically incomplete due to omitted, implicit premises \cite{Hunter2007,Black2012}. For example, the inference from the premise \emph{``the weather report predicts rain''} to the claim \emph{``you should take an umbrella''} hinges on an unstated rule such as \emph{``if the weather report predicts rain, one should take an umbrella''}. Without recovering these hidden links, any automated reconstruction of the argument graph remains partial and potentially misleading.

% \myred{Existing research has approached the problem of logical argument graph construction from various angles, including deductive argumentation frameworks \cite{doi:10.1080/19462166.2013.869765}. Research specifically on enthymemes has evolved along two primary lines. On one hand, natural language processing (NLP) techniques focus on identifying and interpreting enthymemes in textual data \cite{Habernal.et.al.2018.NAACL.ARCT,singh-etal-2022-irac,9963576,doi:10.1177/19462174251344764}. On the other hand, symbolic methods, often based on abduction, aim to infer missing premises that, when combined with explicit ones, logically entail the claim \cite{10.5555/1619645.1619657,10.1007/978-3-642-23963-2_13,Black2012,Hosseini2014,Xydis2020,KR2022-27,Hunter_2022,ORBELEIVA2023108963,Leiva2025,KR2025-11,ijcai2025p495}. Recent work has begun to bridge this gap by both translating free-text enthymemes into logic and inferring implicit premises \cite{feng2026makingimplicitpremisesexplicit}.}

There are methods based on natural language processing for identifying enthymemes in text \cite{Habernal2018,Singh2022,Wei2023,Sviridova2025}, and there are methods based on formal logic for identifying missing premises in a logical representation of an enthymeme using abduction from a logical knowledge base \cite{Hunter2007,Black2012,Hunter2022,BenNaim2025,David2025}.
There are also methods for identifying missing premises from previous moves in a logical representation of a dialogue \cite{Black2008,Dupin2011,Hosseini2014,Xydis2020,Panisson2022,Leiva2023,Leiva2025}.
However, the crucial task of identifying missing premises from common and commonsense knowledge remains challenging. 

In recent work, we have begun to bridge the gap between argument mining \cite{Lawrence2019} and symbolic methods by translating free-text enthymemes into logic and then using neuro-symbolic reasoning based on sentence embeddings to classify the relationship between a pair of formulae as entailment, contradiction, or neutral \cite{Feng2025}. Subsequently, we used an LLM to generate implicit premises that, when translated into logic, can be used to determine whether an explicit claim follows from an explicit premise using neuro-symbolic reasoning \cite{Feng2026}. 

% the problem of missing premises has been considered in a neuro-symbolic pipeline that uses an LLM to generate implicit premises, and then translates these into logic to determine whether the explicit claim follows from the explicit premises \cite{feng2026makingimplicitpremisesexplicit}.

This paper builds on our prior work (i.e., \cite{Feng2026}) by prompting an LLM to generate implicit premises that logically establish a known entailment or contradiction relationship between an explicit premise and an explicit claim. In other words, given a premise-claim pair and its relation label (entailment, contradiction, or neutral), can we identify the implicit premise that makes the logical connection explicit? Furthermore, this paper uses the Microtext Argumentative Corpus \cite{Peldszus2015}, which is a more challenging dataset than those used in our previous work, for the evaluation. Our results indicate that our pipeline significantly enhances the quality and logical coherence of argument graph reconstruction, providing a crucial step toward scalable and accurate logical analysis of argumentative text.

The remainder of this paper is structured as follows: Section~\ref{section:background} reviews background in logical argumentation and enthymeme resolution. Section~\ref{section:pipeline} details our neuro-symbolic pipeline for generating implicit premises. Section~\ref{section:evaluation} describes our experiments with datasets, and Section~\ref{section:discussion} discusses implications and future work.

%%%%%%%%%%%%%%%%%%%%%%%%%%%%%%%%%%%%%%%%%%%%%%
%%%%%%%%%%%%%%%%%%%%%%%%%%%%%%%%%%%%%%%%%%%%%%
\section{Abstract meaning representation}
%%%\section{Background}
\label{section:background}

This section reviews abstract meaning representation (AMR) and how we can translate it into propositional logic.

%%%%%%%%%%%%%%%%%%%%%%%%%%%%%%%%%%%%%%%%%%%%%%
%%%%%%%%%%%%%%%%%%%%%%%%%%%%%%%%%%%%%%%%%%%%%%
%\subsection{Abstract meaning representation}
%%%%%%%%%%%%%%%%%%%%%%%%%%%%%%%%%%%%%%%%%%%%
%\label{section:amr}

Abstract meaning representation (AMR) is a semantic representation language for representing sentences as rooted, labelled, directed, and acyclic graphs (DAGs) \cite{Banarescu2013}. AMR is intended to assign the same AMR graph to similar sentences, even if they are not identically worded.  Negation is represented via the :polarity relation. For example, Figure \ref{f1} represents ``The boy does not want to go."
The numbers after the instance name (such as want-01 above) denote a particular OntoNotes or PropBank semantic frame  \cite{Kingsbury2002}. These frames have different parameters, but the subject is generally denoted by $\texttt{arg0}$ and the object by \texttt{arg1}. 
The parameters, which draw out the semantic roles of the words in the AMR, include \texttt{location} (e.g. ``France"), \texttt{unit} (e.g. ``kilogrammes"), and \texttt{time} (e.g. ``yesterday"). 

% \myblue{The approach was first introduced by Langkilde and Knight in 1998  \cite{langkilde-knight-1998-generation-exploits} as a derivation from the Penman Sentence Plan Language \cite{kasper-1989-flexible}. In 2013, AMRs re-gained attention due to Banarescu et al.  \cite{banarescu-etal-2013-abstract}, and were introduced into NLP tasks such as machine translation and natural language understanding. The modern (post-2010) AMR\footnote{\url{https://github.com/amrisi/amr-guidelines}} draws on predicate senses and semantic roles from the OntoNotes project  \cite{hovy-etal-2006-ontonotes}. }

\begin{figure}
\footnotesize
    \[
\begin{array}{ll}
\mbox{\tt (w / want-01}
& \mbox{\tt (w / want-01}\\
\hspace{1cm}     \mbox{\tt    :arg0 (b / boy)}
& \hspace{1cm} \mbox{\tt :arg0 (b / boy)}\\
\hspace{1cm}   \mbox{\tt    :arg1 (g / go-01}
& \hspace{1cm} \mbox{\tt :arg1 (g / go-01}\\
\hspace{2cm}  \mbox{\tt          :arg0 b))}
& \hspace{2cm}\mbox{\tt :arg0 b}\\
&\hspace{2cm}\mbox{\tt  :polarity -))}
    \end{array}
    \]
    %\vspace{-5mm}
    \caption{AMR for the sentence ``The boy wants to go." (left) and ``The boy does not want to go." (right).\\}
    \label{f1}
\end{figure}

% \myblue{AMR uses the same structure to represent semantically similar texts by making several simplifying assumptions. AMR cannot represent verb tenses nor distinguish between verbs and nouns. Also it does not represent articles, quote marks, or the singular and plural.
% Nonetheless, AMR offers a valuable formal abstraction of the meaning of sentences.}

% AMR may be shown as a graph or a tree, where the tree corresponds exactly to the AMR text format  \cite{goodman-2020-penman}. Although the graph representation can always be generated from the tree representation, the opposite is not always possible because there can be more than one tree that can map to the same graph \cite{chanin2023neuro}.

% \begin{figure}
% \centering
%     \footnotesize
%     \begin{verbatim}
%     (w / want-01
%         arg0 (b / boy)
%         arg1 (g / go-01
%             arg0 b
%             :polarity -))
%     \end{verbatim}
%     \vspace{-5mm}
%     \caption{AMR for the sentence ``The boy does not want to go."}
%     \label{fig:basicamr2}
% \end{figure}

\paragraph{Text-to-AMR Parser} In our pipeline, we use the IBM Transition AMR parser\footnote{\url{https://github.com/IBM/transition-amr-parser/tree/master}} to load the pre-trained ensemble AMR 3.0 model (AMR3-joint-ontowiki-seed43), which combines smatch-based ensembling techniques with ensemble distillation  \cite{Lee2021} to translate each sentence of text into an AMR graph.

%%%%%%%%%%%%%%%%%%%%%%%%%%%%%%%%%%%%%%%%%%%%%%
%%%%%%%%%%%%%%%%%%%%%%%%%%%%%%%%%%%%%%%%%%%%%%
%\subsection{Translation of AMR into Logic}
%%%%%%%%%%%%%%%%%%%%%%%%%%%%%%%%
%\label{section:automatedreasoning}

\paragraph{AMR-to-Propositional-Logic Translator} An advantage of AMR is that we can easily transform an AMR graph into first-order logic formulas using the Bos algorithm \cite{Bos2016}. 
% \myblue{which translates each graph into a nested conjunction of atoms, where each monadic atom is a concept and each dyadic atom is relation between a pair of concepts  \cite{chanin2023neuro}. 
% An example of this is shown below, where the AMR for ``The boy does not want to go" from Figure \ref{f1} is converted into a logical formula as follows.
% \[
% \begin{array}{l}
% \tt \exists w ( \exists b(want(w) \wedge arg0(w, b) \wedge boy(b) %\\
% %%%\hspace{2cm}   
% \tt \wedge \hspace{1mm} \neg \exists g (arg1(w, g) \wedge go(g) \wedge arg0(g, b))))
% \end{array}
% \]
% }
There is an open-source Python library based on the Bos algorithm, the AMR-to-logic converter \cite{Chanin2023}, for translating AMR graphs into first-order logic. We extended this library to create our AMR-to-propositional-logic translator by rewriting each first-order logic formula as a propositional logic formula and grounding out the existentially quantified variables with new constants. We refer to each such propositional formula as an AMR formula.
% \myblue{For this, we make the assumption that for each existentially quantified variable, there is a specific entity that can represent the quantified variable. This can be viewed as a Skolem constant.
% We choose each constant symbol for this grounding as follows: For each monadic predicate $r(a)$, we use $r$ as the constant symbol to replace the variable symbol $a$ throughout the formula. 
% This then means the monadic predicates are now redundant and so we delete them.}
%%%%Furthermore, for our choice of constant symbol, we use the predicate symbol of the monadic predicate that 
% For the above example, the  propositional logic formula is as follows:
% \[
% \begin{array}{l}
% \tt want(w) \wedge arg0(w, b) \wedge boy(b) \\
% \hspace{2cm}     \tt \wedge \hspace{1mm}  \neg (arg1(w, g) \wedge go(g) \wedge arg0(g, b))
% \end{array}
% \]
% \tony{Besides the above formula, we also performed extra processing. We merged the monadic predicates into dyadic predicates as a simplification. }
For the above example, the  propositional logic formula is simplified as follows.
\[
\begin{array}{l}
\tt arg0(want, boy)  \tt \wedge \hspace{1mm}  \neg (arg1(want, go)\wedge arg0(go, boy))
\end{array}
\]
% \new{Moreover, we merged predicates like "mode" and "quant" For example,}
% \[
% \begin{array}{l}
% \tt arg0(run, boy) \wedge mod(boy, happy) = \tt \tt arg0(run, happy\ boy)
% \end{array}
% \]
% \[
% \begin{array}{l}
% \tt arg0(run, boy) \wedge quant(boy, 2) = \tt \tt arg0(run, 2\ boy)
% \end{array}
% \]

For our pipeline, we assume the usual definitions for propositional logic.
We start with a set of propositional atoms (letters), and we construct formulas in the usual way using the connectives for negation $\neg$, conjunction $\wedge$, disjunction $\lor$, implication $\rightarrow$, and biconditional $\leftrightarrow$.

%%%%%%%%%%%%%%%%%%%%%%%%%%%%%%%%%%%%%%%%%%%%%%
%%%%%%%%%%%%%%%%%%%%%%%%%%%%%%%%%%%%%%%%%%%%%%
\section{Pipeline}
\label{section:pipeline}
%%%%%%%%%%%%%%%%%%%

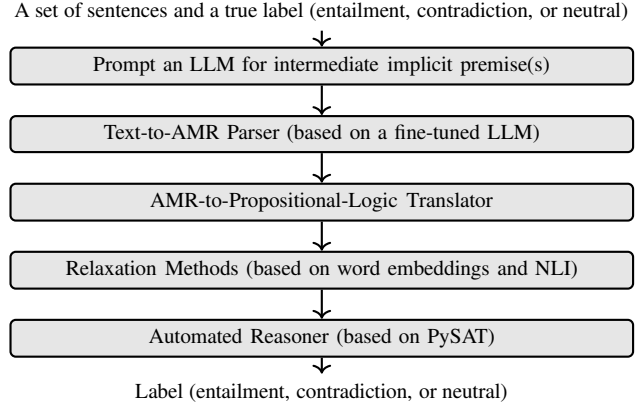
\begin{figure}[t]
\begin{center}
\begin{tikzpicture}[->,thick,scale=0.9,
ARG/.style={draw,text centered, text width=80mm,
shape=rectangle, 
rounded corners=2pt,
fill=gray!20,font=\footnotesize}]
%\node[draw,rectangle,fill=yellow] (n1)  at (4.5,4.5) {\scriptsize $n_1$};
\node[] (a0)  at (0,7.8) {\footnotesize A set of sentences and a true label (entailment, contradiction, or neutral)};
\node[ARG] (a00)  at (0,7) {\footnotesize Prompt an LLM for intermediate implicit premise(s)};
\node[ARG] (a1)  at (0,6) {\footnotesize Text-to-AMR Parser (based on a fine-tuned LLM)};
\node[ARG] (a2)  at (0,5) {\footnotesize AMR-to-Propositional-Logic Translator};
\node[ARG] (a3)  at (0,4) {\footnotesize Relaxation Methods (based on word embeddings and NLI)};
\node[ARG] (a4)  at (0,3) {\footnotesize Automated Reasoner (based on PySAT)};
\node[] (a5)  at (0,2.2) {\footnotesize Label (entailment, contradiction, or neutral)};
\path (a0)[] edge[] (a00);
\path (a00)[] edge[] (a1);
\path (a1)[] edge[] (a2);
\path (a2)[] edge[] (a3);
\path (a3)[] edge[] (a4);
\path (a4)[] edge[] (a5);
% \path (a5)[] edge[] (a6);
\end{tikzpicture}
\end{center}
\caption{Our neuro-symbolic pipeline where the input is a set of natural language sentences (e.g., a premise, implicit premises, and a claim), and the output is a label.}
\label{fig:pipeline}
\end{figure}
%%%%%%%%%%%%%%%%

Our neuro-symbolic pipeline\footnote{\url{https://github.com/fxy-1117/COMMA2026}},
which is summarized in Figure~\ref{fig:pipeline}, consists of five main components:
An LLM to generate  intermediate implicit premises (Section \ref{llmg}); 
A text-to-AMR parser (Section \ref{section:background}); %\ref{section:amr}); 
An AMR-to-propositional-logic translator (Section \ref{section:background}); %\ref{section:automatedreasoning}); 
A set of methods for relaxing the propositional formulas (Sections \ref{section:similarity} 
and \ref{section:translate});
And an automated reasoner based on PySAT (Section \ref{section:classification}). 
%%%This pipeline is summarized in Figure \ref{fig:pipeline}.

% In the rest of this section, we explain how we obtain intermediate premises, how we represent the AMR formulas that come from the AMR-to-propositional-logic translator, how we rewrite AMR formulas to abstract formulas for use by a SAT solver, and how we use relaxation methods to simplify these abstract formulas when determining whether entailment holds between pairs of formulas. 

%%%%%%%%%%%%%%%%%%%%%%%%%%%%%%%%%%%%%%%%%%%%%%
%%%%%%%%%%%%%%%%%%%%%%%%%%%%%%%%%%%%%%%%%%%%%%
\subsection{Generate implicit premises}
\label{llmg}
%%%%%%%%%%%%%%%%%%%%%%%%%%%%%%%%%%%%%%%%%%%%%%

We employ a large language model (LLM) to generate implicit premises by providing it with an explicit premise, a claim, and the label indicating whether the claim follows from or contradicts the premise. The model is then prompted to produce either one or six intermediate reasoning steps that make explicit how the claim is derived from or contradicts the premise. These intermediate steps form a chain of reasoning, as explored in \cite{Wei2022}, which incrementally bridges the gap between the explicit premise and claim. 

As the number of steps increases, the reasoning exposes finer logical connections and provides additional detail on how the claim either follows from or contradicts the premise. 
%%%%%%We not only generate helpful steps that support the premise in entailing the claim, but we also generate unhelpful steps that lead the premise to contradict or be neutral with respect to the claim, in order to evaluate our pipeline. 
%%%%In addition to the one-step implicit premises provided by the two datasets, 
%%%we also generate extra one-step, two-step, and three-step implicit premises. 
The LLM used is DeepSeek v3.2 \cite{Deepseek2025}, and our prompt is presented with the code. The implicit premises, like the explicit premises and claims, are natural language statements, and so we use the text-to-AMR parser and the AMR-to-propositional-logic translator to obtain logical formulas which we handle in Sections \ref{section:similarity} and \ref{section:translate}. 

% For the evaluation (Section \ref{section:evaluation}), the implicit premises generated above are helpful premises. We used the same method to also generate unhelpful premises.  So for a premise and claim, the unhelpful premise together with the premise would contradict or be neutral with respect to the claim. So for the evaluation, we had one-, two-, and three-step helpful premises and one-, two-, and three-step unhelpful premises. 

%%%%%%%%%%%%%%%%%%%%%%%%%%%%%%%%%%%%%%%%%%%%%%
%%%%%%%%%%%%%%%%%%%%%%%%%%%%%%%%%%%%%%%%%%%%%%
\subsection{Representing formulas}
%%%%%%%%%%%%%%%%%%%%%%%%%%%%%%%%%%%%%%%%%%%%%%

An AMR formula is composed of a set of atoms together with the $\neg$ and $\wedge$ logical operators.
Since each atom is ground, an AMR formula is a propositional logic formula.
Let ${\cal A}$ be a set of AMR atoms (i.e. ground dyadic predicates of the form $r(a,b)$ where $a$ and $b$ are constant symbols). 
The set of {\bf AMR formulas}, denoted ${\cal L}$, is defined inductively as follows:
If $\alpha\in{\cal A}$, then $\alpha\in{\cal L}$;
If $\alpha,\beta\in{\cal L}$, then $\alpha\wedge\beta\in{\cal L}$;
And if $\alpha\in{\cal L}$, then $\neg\alpha\in{\cal L}$.

Given a natural language sentence $S$, our AMR-to-propositional-logic translator identifies an AMR formula $\phi$ 
that {\bf represents} $S$. 

\begin{example}
\label{ex:amrformula}
The AMR formula $\tt \neg arg0(go,car)$
represents the sentence ``the car does not go".
\end{example}

An abstract formula is also a propositional formula. 
% It is composed from a set of propositional letters together with the $\neg$ and $\wedge$ logical operators.
% The set of abstract formulas is also a subset of the set of propositional formulas. 
Let ${\cal P}$ be a set of propositional letters. 
The set of {\bf abstract formulas}, denoted ${\cal F}$, is defined inductively as follows:
If $\alpha\in{\cal P}$, then $\alpha\in{\cal F}$;
If $\alpha,\beta\in{\cal F}$, then $\alpha\wedge\beta\in{\cal F}$;
And if $\alpha\in{\cal F}$, then $\neg\alpha\in{\cal F}$.

If $|{\cal A}| = |{\cal P}|$, then for each $\alpha \in {\cal L}$, there is a $\beta\in {\cal F}$ (and vice versa) such that $\alpha$ and $\beta$ are isomorphic (i.e. they have the same syntax tree except for the atoms associated with the leaves). For example, $\tt \neg x_1$ is an abstract formula that is isomorphic to the AMR formula in Example \ref{ex:amrformula}.

Given an AMR formula or an abstract formula, denoted $\phi$, let ${\sf Atoms}(\phi)$ denote the set of atoms used in $\phi$.

%%%%%%%%%%%%%%%%%%%%%%%%%%%%%%%%%%%%%%%%%%%%%%
%%%%%%%%%%%%%%%%%%%%%%%%%%%%%%%%%%%%%%%%%%%%%%
\subsection{Similarity measures}
\label{section:similarity}
%%%%%%%%%%%%%%%%%%%%%%%%%%%%%%%%%%%%%%%%%%%%%%

% In order to consider how we can translate AMR formulas into abstract formulas, 
% we need to consider when two atoms can be regarded as equivalent, and therefore mapped to the same propositional letter. 

In the next two subsections, we explain how we translate each AMR formula into an abstract formula, 
which we then use with the PySAT automated reasoner.
For the translation, we check whether two atoms in an AMR formula can be regarded as equivalent, and therefore mapped to the same propositional letter in the corresponding abstract formula.
For this, if for two AMR atoms $\alpha$ and $\beta$, 
the similarity between the two strings of words corresponding to the two atoms is greater 
than a threshold $\tau_m$ (as explained below), 
then we treat them as equivalent (denoted $\alpha\simeq\beta$) in the corresponding abstract formula,  
and so it is a relaxation of the AMR formula.
%and so constitutes a form of neuro-symbolic reasoning.}

An {\bf embedding} of a string of words
is a vector $v$ obtained by an injective mapping function $\embed$. 
In our pipeline, the $\embed$ function is the sentence transformer BAAI general embedding model bge-small-en-v1.5 \cite{Xiao2024} (alternatives can easily be used) that encodes a string of words as a high-dimensional vector. 

%%\begin{definition}
The {\bf similarity} between two embedding vectors $v_1$ and $v_2$ is defined as follows, where $\theta$ is the angle between the vectors, $v_1 \cdot v_2$ is the dot product of $v_1$ and $v_2$, and $||v_1||$ represents the L2 norm.
\[
{\similar}(v_1,v_2) = cos(\theta) = \frac{v_1 \cdot v_2}{||v_1||||v_2||}
\]
%%\end{definition}

To obtain a string of words corresponding to an AMR atom, we created a set of 29 templates.
Each template is based on taking the information in an AMR atom in an AMR formula
and representing that information as a natural language sentence.
Furthermore, each template is designed for a type of AMR atom (i.e. for the predicate name of the atom). 
For example, for the AMR atom $\tt arg0(play,man)$,
the predicate name is $\tt arg0$, and so we can use a template for $\tt arg0$
such as ``[Y] is the agent performing action [X]" which is a string where [X] and [Y] are placeholders for the first and second arguments of the atom (i.e. $\tt play$ and $\tt man$ in this example).
So from this template, we can instantiate it using the dyadic atom to get ``{\em man is the agent performing action play}" as the natural language sentence to be used for the word embedding. In order to do this, we require the following definition.

% So from this template, we get ``man is the agent performing action play." as the natural language sentence used as the prompt to an LLM. In order to do this, we require the following definition.

\begin{definition}
Let $\phi$ be an AMR formula,
and let $r(a,b)$ be an AMR atom in $\phi$ (i.e. $r(a,b) \in {\sf Atoms}(\phi)$). 
Also let $T$ be a template for $r$ with placeholders $[{\rm X}]$ and $[{\rm Y}]$. 
The {\bf instantiate function}, denoted $\prompt$, is defined as follows: 
$\prompt(r(a,b), T)$ = $I$, where $I$ is the instantiation of $T$ in which $[{\rm X}]$ is replaced by $a$ and $[{\rm Y}]$ is replaced by $b$.
\end{definition}

Some examples of templates are given below, with the AMR predicate name.

\begin{itemize}
\item ($\tt purpose$) ``[Y] is the purpose of action [X]."  
\item ($\tt time$) ``[Y] is when action [X] takes place."  
\item ($\tt arg1$) ``[Y] is the object involved in action [X]."
\end{itemize}

Next, we use the embedding function $\embed$ to obtain the sentence embeddings of the instantiations of templates.

\begin{example}
\label{e13}
Let $T$ = {\rm ``[Y] is the agent performing action [X]."} be the template for the AMR predicate name $\tt arg0$. 
Therefore, for the AMR atom $\tt arg0(play,child)$,
$\prompt({\tt arg0(play,child)}, T)$ = ``child is the agent performing action play."
\end{example}

% \begin{example}
% \label{e13}
% \tony{For the binary atom $\tt domain(red,car)$, the AMR predicate name is $\tt domain$.
% Let $S$ be the template $\tt domain$ where "[T2] is the domain of action [T1]". Thus, ${\sf Replace}({\tt domain(red,car)}, \\
%  "[T2]\ is\ the\ domain\ of\ action\ [T1]")$ = "car is the domain of action red"}
% \end{example}

% \begin{example}
% \label{e13}
% \tony{Here are some templates for other AMR predicate title, $\tt mod([T1],[T2])$: ``[T2] modifies action [T1].", $\tt arg0([T1],[T2])$: ``[T2] is the agent performing action [T1].", $\tt arg1([T1],[T2])$: ``[T2] is the object involved in action [T1].".}
% \end{example}

For the following definition of a matching relation, we assume that one AMR formula refers to a premise and the other refers to a claim.

\begin{definition}
Let an AMR formula $\phi$ be a premise and an AMR formula $\psi$ be a claim,
let $\alpha\in {\sf Atoms}(\psi)$ be an AMR atom of the form $r(a,b)$, 
let $T_1$ be a template for $r$, 
let $\beta\in {\sf Atoms}(\phi)$ be an AMR atom of the form $q(c,d)$,
let $T_2$ be a template for $q$, 
let $\tau_m \in [0,1]$ be the {\bf neuro-matching threshold}, 
and let $\embed$ be an embedding function.
%and let $h$ be a function such that $h(r(a,b)) = {\sf Replace}(r(a,b), S)$ if $\gamma$ is dyadic atom and we have a template $S$ for $r$.
The {\bf neuro-matching relation}, denoted $\simeq$, is defined as follows,
where ${\sf Near}(\alpha)$ = $\{ (\beta',x) \mid \beta'\in {\sf Atoms}(\phi)$ and 
$x > \tau_m$
and ${\similar}({\embed}(\prompt(\alpha,T_1)),{\embed}(\prompt(\beta',T_2))) = x\}$.
\[
\begin{array}{l}
\alpha \simeq \beta \mbox{ iff }  (\beta,x) \in {\sf Near}(\alpha) \mbox{ and }\forall (\beta',y) \in {\sf Near}(\alpha), x \geq y
\end{array}
\]
%%%where $.
% where ${\sf Sim}(\alpha)$ = $\{ (\beta,x) \mid \beta\in {\sf Atoms}(\phi)$ and $x > \tau_m$
% and ${\sf similarity}({ f}(h(\alpha)),{ f}(h(\beta))) = x\}$ 
\end{definition}

This definition finds the best premise match above the threshold for each claim atom. Thus, if the premise atoms are $\beta_1,\ldots,\beta_n$ with scores $s_1,\ldots,s_n$, then $\alpha\simeq\beta$ holds when $\beta=\beta_i$ and $s_i=\max(s_1,\ldots,s_n)$.
\begin{example}
\label{example:neuromatch}
Let the text for the premise be ``A tiger is walking in the cage.",
and let the text for the claim be ``A tiger is moving."
We have the following AMR formulas.
\[
\begin{array}{c}
\tt arg0(walk, tiger) \wedge location(walk, cage)\\
  \tt arg0(move, tiger) 
\end{array}
\]
So $\tt  arg0(walk, tiger) \simeq \tt arg0(move, tiger)$,
with template $T$ for $\tt  arg0$,
the following similarity,
and $\tau_m = 0.6$.
\[
\begin{array}{l}
    {\similar}(\embed(\prompt({\tt arg0(walk, tiger)},T)),\embed(\prompt({\tt arg0(move, tiger)},T)))  = 0.8483
\end{array}
\]

\end{example}
After identifying the best match for an AMR atom in a claim, we perform a contradiction check on this matched pair using a Natural Language Inference (NLI) model\footnote{mDeBERTa-v3-base-xnli-multilingual-nli-2mil7} \cite{Laurer2023}. The NLI model evaluates a pair of sentences \((S_1, S_2)\) by identifying three scores, each in the [0,100] interval: 
\(s_{\text{Ent}}(S_1, S_2)\) is the degree to which \(S_2\) is entailed by \(S_1\); 
\(s_{\text{Con}}(S_1, S_2)\) is the degree of conflict between \(S_1\) and \(S_2\);
and \(s_{\text{Neu}}(S_1, S_2)\) is the degree to which \(S_1\) and \(S_2\) are unrelated.
The model returns the label with the highest score (with a random choice in case of a tie).

\begin{definition}
% \myred{Let \(\mathcal{R}\) be the set of all possible premises (evidence),
% \(\mathcal{C}\) the set of all possible hypotheses (claims)},
Let \(\mathcal{S}\) be the set of all natural language sentences,
and let $\mathcal{C} = \{\text{Ent}, \text{Con}, \text{Neu}\}$
be the set of outcomes (entailment, contradiction, and neutrality). 
A {\bf natural language inference (NLI) function}
$\nli : \mathcal{S} \times \mathcal{S} \rightarrow \mathcal{C}$
is defined as follows, with scores $s_{\text{Ent}}$, $s_{\text{Con}}$, and $s_{\text{Neu}}$
for a given pair \((S_1, S_2)\).
\[
\nli(S_1,S_2) = \arg\max_{\sigma \in \mathcal{C}} s_{\sigma}(S_1, S_2)
\]
\end{definition}

%%%So the NLI function selects the label corresponding to the maximum score. 

% The predicted relationship \(\hat{\ell}\) is determined by selecting the label corresponding to the maximum score:
% \[
% \hat{\ell} = \arg\max_{\ell \in \mathcal{S}} s_\ell(R, C)
% \]
% \end{definition}

\begin{definition}
Let an AMR formula $\phi$ be a premise and an AMR formula $\psi$ be a claim,
let $\alpha\in {\sf Atoms}(\psi)$ be an AMR atom of the form $r(a,b)$, 
let $T_1$ be a template for $r$, 
let $\beta\in {\sf Atoms}(\phi)$ be an AMR atom of the form $q(c,d)$,
let $T_2$ be a template for $q$, let $\nli$ be an NLI function,
and let $\tau_c \in [0,100]$ be the {\bf neuro-contradict threshold}. The {\bf neuro-contradict relation} $\perp$
%and let $h$ be a function such that $h(r(a,b)) = {\sf Replace}(r(a,b), S)$ if $\gamma$ is dyadic atom and we have a template $S$ for $r$.
is defined as follows 
% \[
% \begin{array}{ll}
% \alpha \perp \beta\ \mbox{iff}
% & \mathsf{N}(\prompt(\alpha,T_1),\prompt(\beta,T_2)) = \text{Con}\\
% & \hspace{10mm} \mbox{ and } s_{\text{Con}}(\prompt(\alpha,T_1),\prompt(\beta,T_2)) \geq \tau_c
% \end{array}
% \]
\[
\begin{array}{l}
\alpha \perp \beta\ \mbox{ iff }
\nli(\prompt(\alpha,T_1),\prompt(\beta,T_2)) = \text{Con}
\mbox{ and } s_{\text{Con}}(\prompt(\alpha,T_1),\prompt(\beta,T_2)) \geq \tau_c
\end{array}
\]

\end{definition}
\begin{example}
\label{example:neuromatchc}
Let the text for the premise be ``A tiger is walking in the cage.",
and let the text for the claim be ``The tiger is sleeping in the cage."
We have the AMR formulas.
\[
\begin{array}{c}
\tt arg0(walk, tiger) \wedge \tt location(walk, cage)\\
  \tt arg0(sleep, tiger) \wedge \tt location(sleep, cage)
\end{array}
\]
When $\tau_c = 80$, $\tt arg0(walk, tiger) \perp \tt arg0(sleep, tiger)$, 
and $\tt location(walk, cage) \!\perp\! \tt location(sleep, cage)$
hold, given the following outputs for templates $T_1$ and $T_2$ for $\tt arg0$ and $\tt location$, respectively.
\[
\begin{array}{c}
    {\nli}( \prompt({\tt arg0(walk, tiger)},T_1),
    \prompt({\tt arg0(sleep, tiger)},T_1)) = \text{Con}\\
    s_{\text{Con}}( \prompt({\tt arg0(walk, tiger)},T_1),
   \prompt({\tt arg0(sleep, tiger)},T_1)) = 85\\
\\
% \end{array}
% \]
% \[
% \begin{array}{c}
   {\nli}( \prompt({\tt location(walk, cage)},T_2),
 \prompt({\tt location(sleep, cage)},T_2)) = \text{Con}\\
    s_{\text{Con}}(  \prompt({\tt location(walk, cage)},T_2),
 \prompt({\tt location(sleep, cage)},T_2)) = 82
\end{array}
\]

\end{example}

The $\simeq$ relation is reflexive and symmetric, whereas the $\perp$ relation is irreflexive and symmetric.

%%%%%%%%%%%%%%%%%%%%%%%%%%%%%%%%%%%%%%%%%%%%%%
%%%%%%%%%%%%%%%%%%%%%%%%%%%%%%%%%%%%%%%%%%%%%%
\subsection{Translating AMR into abstract formulas}
%%%%%%%%%%%%%%%%%%%%%%%%%%%%%%%%%%%%%%%%%%%%%%
\label{section:translate}

We now consider how we can translate each AMR formula into an abstract formula.
The first aim is to represent each AMR atom by an abstract atom of the form $\tt x_i$ in order to facilitate use by a PySAT solver, and the second aim is to take advantage of the neuro-matching and neuro-contradict relations to simplify and constrain the abstract formula. 
For example, if we have a premise $\tt x_1 \wedge x_2 \wedge x_3$ and a claim $\tt x_1 \wedge x_4$ and $\tt x_3$ and $\tt x_4$ are very similar concepts, then we can change the claim to $\tt x_1 \wedge x_3$, and then show entailment holds using the CNF versions of these relaxed formulas with PySAT. 
Similarly, if we have a premise $\tt x_5 \wedge x_6$ and a claim $\tt x_7$, 
and the $\tt x_5 \bot x_7$ relationship holds, 
then we can change the claim to $\tt \neg x_5$.

\begin{definition}
\label{def:translation}
Given a set of AMR formulas $\Phi$, 
and a set of neuro-matching and neuro-contradict relationships $\Psi$, 
the function $g: {\cal A} \rightarrow {\cal P} \cup \{ \neg x \mid x \in {\cal P} \}$ is a {\bf mapping} for $\Phi$ and $\Psi$
iff for all $\phi,\phi'\in\Phi$, 
for all $\alpha\in {\sf Atoms}(\phi)$, 
for all $\beta\in {\sf Atoms}(\phi')$,
(1) $\alpha \simeq \beta\in\Psi  \mbox{ iff } g(\alpha) = g(\beta)$;
and (2) $\alpha \perp \beta\in\Psi  \mbox{ iff } g(\alpha) = \neg g(\beta)$.
For the tautology $\top$, $g(\top) = \top$.
\end{definition}

The above definition ensures that if there are similar atoms, according to $\simeq$, 
(resp. contradictory atoms, according to $\bot$), 
then they are translated to the same atom (resp. complementary literals) in the abstract formulas.
% If we use standard matching $=$ for $\simeq$, then a bijection from ${\cal A}$ to ${\cal P}$ is an example of a translation function. 

\begin{example}
\label{ex:translation}
For $\phi_1 = {\tt arg1(car,red)} \wedge {\tt arg2(car,fast)}$, 
$\phi_2 = {\tt arg1(car,red)}$, 
and $\Phi = \{\phi_1,\phi_2\}$,
a mapping $g$ is 
$g({\tt  arg1(car,red)}) = {\tt x_1}$
and $g({\tt arg2(car,fast)}) = {\tt x_2}$.
\end{example}

\begin{example}
\label{e7}
Continuing Ex~\ref{example:neuromatch},
%%%%a mapping $g$ is 
let $g$ be a mapping $g({\tt location(walk, cage)}) = \tt x_2$
and $g({\tt arg0(walk, tiger)}) = g(\tt arg0(move, tiger)) = \tt x_1$.
\end{example}

\begin{example}
\label{e8}
Continuing Ex~\ref{example:neuromatchc},
$g({\tt arg0(sleep,tiger)})=\neg g({\tt arg0(walk,tiger)})=\neg{\tt x_1}$ and
$g({\tt location(sleep,cage)})=\neg g({\tt location(walk,cage)})=\neg{\tt x_2}$.
\end{example}

Next, we specify how an AMR formula is translated into an abstract formula using a mapping function.

\begin{definition}
Let $g$ be a mapping for a set of AMR formulas $\Phi$ and a set of neuro-matching and neuro-contradict relationships $\Psi$.
For $\phi \in \Phi$, a {\bf translation} of $\phi$ is
$\myabstract_g(\phi)$ where $\myabstract_g$ is defined as:
(1) $\myabstract_g(\alpha\wedge\beta) = \myabstract_g(\alpha) \wedge \myabstract_g(\beta)$;
(2) $\myabstract_g(\neg\alpha) = \neg\myabstract_g(\alpha)$;
and (3) $\myabstract_g(\alpha) = g(\alpha)$ when $\alpha\in{\cal A}$.
For $\Phi$, let $\myabstract_g(\Phi)$ = $\{\myabstract_g(\phi) \mid \phi \in \Phi\}$.
\end{definition}

\begin{example}
Continuing Ex \ref{ex:translation},
a translation of $\phi_1$ is ${\tt x_1} \wedge {\tt x_2}$
and a translation of $\phi_2$ is ${\tt x_1}$.
\end{example}

The use of embeddings and NLI allows for the identification of neuro-matching and neuro-contradict relations, which can then be used to rewrite AMR formulas into abstract formulas, where the latter are relaxations of the former. 
So our relaxation methods rewrite the AMR formulas (output from the AMR-to-propositional-logic translator) into abstract formulas for use in automated reasoning as described next.

%%%%%These abstract formulas are then rewritten into CNF, and tested for entailment using a PySAT solver (as explained in Section \ref{section:automatedreasoning}). 

%%%%%%%%%%%%%%%%%%%%%%%%%%%%%%%%%%%%%%%%%%%%%%
%%%%%%%%%%%%%%%%%%%%%%%%%%%%%%%%%%%%%%%%%%%%%%
\subsection{Automated Reasoning}
\label{section:classification}
%%%%%%%%%%%%%%%%%%%%%%%%%%%%%%%%%%%%%%%%%

For the automated reasoning, we transform all the abstract formulas (from the previous section) into conjunctive normal form (CNF) using SymPy \cite{Meurer2017}.
% CNF is defined in the usual way: (1) A literal is either a propositional variable, or the negation of one; (2) A clause is a disjunction of literals; and (3) A formula in conjunctive normal form (CNF) if it is a literal, or a clause, or a conjunction of clauses. 
Then, we use PySAT \cite{Ignatiev2018}, which integrates several widely used state-of-the-art SAT solvers as theorem provers to check whether a CNF is consistent. 
%%%return True if consistent or otherwise False. 
%%%%An input to PySAT will be CNF but represented as a list which consists of lists and integers. 
%%Finding an interpretation for a Boolean formula is known as the Boolean satisfiability problem (SAT). 
% In general, the question is to find whether the variables in a specific Boolean formula can be consistently substituted with the values TRUE or FALSE in a manner that results in evaluating the formula TRUE. In this case, the formula is said to be satisfiable. Conversely, if there is no such assignment,  the function described by it is FALSE for all potential variable assignments, and the formula is unsatisfiable. 
%%%For example, the formula $\tt a \wedge \neg b$ is satisfiable as $\tt a$ = TRUE and $\tt b$ = FALSE resulting in the formula being TRUE. In contrast, $\tt a \wedge \neg a$ is unsatisfiable. 
% \begin{example}
%     The CNF $ \tt x_1 \wedge x_2 \wedge  (\neg x_1 \lor \neg x_2)$ will be represent as [1,2,[-1,-2]]
% \end{example}
The aim of our pipeline is to identify the relationship between a premise $\phi$ (which may be a conjunction of an explicit premise and one or more intermediate premises) and a claim $\psi$. Proving that $\phi$ entails $\psi$ is equivalent to determining whether $\phi \land\neg \psi$ is inconsistent.
%%%%%\myblue{To do this, we need to change $\phi \land\neg \psi$ into a CNF formula, and then we can directly use PySAT to check consistency.} 
Similarly, to prove whether $\phi$ contradicts $\psi$, we need to determine whether $\phi\wedge\psi$ is inconsistent. 
%%%%\myblue{To do this, we need to change $\phi\land\psi$ into a CNF formula, and then we can directly use PySAT to check consistency.}

%%%%%%%%%%%%%%%%%%%%%%%%%%%%%%%%%%%%%%%%%%%%%%
%%%%%%%%%%%%%%%%%%%%%%%%%%%%%%%%%%%%%%%%%%%%%%
\section{Experiments}
\label{section:evaluation}
%%%%%%%%%%%%%%%%%%%%%%%%%%%%%%%%%
To evaluate our pipeline, we used the Microtext Argumentative Corpus,
which we describe next, in three experiments.

\begin{figure}[htbp]
    \centering
    % Ensure \usepackage{graphicx} is included in your document preamble
    %\resizebox{\columnwidth}{!}{%
    %\resizebox{scale=0.8, transform shape}{!}{
        \begin{tikzpicture}[scale=0.72, transform shape,
            every node/.style={rectangle, draw, fill=gray!10,text width=5cm, align=left, minimum height=1cm, inner sep=5pt},
            support/.style={->, thick, solid},
            rebut/.style={->, thick, dashed},
            undercut/.style={->, thick, dotted}
        ]
        
        \node at (0,0)    (e1) {1: I think it's possible for older people to be better parents.};
        \node at (0,4.5)  (e2) {2: They have had time to mature more and learn about themselves.};
        \node at (6,1.5)  (e4) {4: so they can provide for a child.};
        \node at (6,4.5)  (e3) {3: They're more likely to have steady careers going.};
        \node at (-6,1.5) (e5) {5: Some might say being too old risks the child being disabled.};
        \node at (-6,4) (e6) {6: but insisting all babies be perfect is demanding the impossible; 7: as well as demeaning to people already living who are disabled.};

        % Draw edges based on relationships
        \draw[support]  (e2) to node[draw=none, fill=none, right] {support} (e1);
        %\draw[support]  (e4) to (e1);
        \draw[support]  (e3) to node[draw=none, fill=none, right] {support}  (e4);
        %\draw[rebut]    (e5) to (e1);
        \draw[support] (e6) to node[draw=none, fill=none, right] {undercut} (e5);
        \draw[support] (e5) |- (e1);
        \draw[support] (e4) |- (e1);
        \node[draw=none, fill=none] at (-3.3,0.5) (n1) {rebuttal};
        \node[draw=none, fill=none] at (8.7,0.5) (n2) {support};
        \end{tikzpicture}%
    %}
    \caption{Argument graph example (text number 008) where 
    %%the solid line represents support relation, the dashed line represents rebuttal relation, and the dotted line represents undercut relation. 
    Segment 1 is the central claim. It is directly supported by Segment 2 and Segment 4. Segment 3, in turn, provides support for Segment 4. The claim is challenged by Segment 5, which rebuts it, but this rebuttal is itself undercut by Segments 6 and 7.}
    \label{fig:ag}
\end{figure}
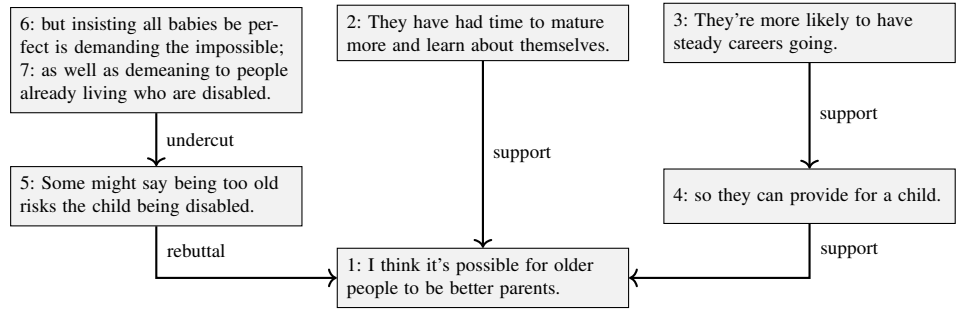

\paragraph{Method} The Microtext Argumentative Corpus~\cite{Peldszus2015} is a  resource specifically designed for argument mining research. The corpus consists of 112 short, self-contained argumentative texts, split into 576  segments, which have been manually annotated by expert annotators to identify argument components (viz.,
a central claim, claims, and premises) and the support or attack relationships (rebuttal and undercut) between them. 
%%%%This fine-grained, discourse-level annotation provides a dataset for training and evaluating models on argumentation tasks such as component identification and relation classification. 
Figure \ref{fig:ag} is an argument graph of an example from the corpus.

To evaluate our pipeline, we translated the Microtext dataset into a three-class 
%%%%%Natural Language Inference (NLI) 
dataset by generating premise-claim pairs from argument segments: 
From the argument graph representation (of an argumentative text in the Microtext dataset), each arc gives a premise-claim pair where the segment in the source (resp. target) node is the premise (resp. claim).
The label for each premise-claim pair is assigned using the original Microtext dataset: \textbf{Entailment} ($ent$) for a \textit{support} relation from premise to claim; \textbf{Contradiction} ($con$) for a \textit{rebuttal} or \textit{undercut} relation from premise to claim; and \textbf{Neutral} ($neu$) for segments $S_1$ and $S_2$ with no relationship between them and with $\nli(S_1,S_2) = Neu$ with confidence $s_{Neu}(S_1,S_2) \geq 99$ (recall that $\nli$ is the NLI function).

\begin{example}
For the argument graph in Figure~\ref{fig:ag},
we get the following labelling: 
the Entailment label for $(2,1)$, $(4,1)$, and $(3,4)$;
the Contradiction label for $(5,1)$ and $(6\&7,5)$, where $\&$ denotes a conjunction of segments;
and the Neutral label for $(3,5)$ and $(3,1)$.
% With this encoding as labelled premise-claim pairs, we see that a segment  can be a claim in one relationship and a premise in another (e.g. $4$ or $5$).
% Furthermore, a premise can undercut or rebut a claim (e.g. $6\&7$ undercuts $5$
% For example, $4$ is 
\end{example}

% ((2,1),Entailment), 
% ((4,1),Entailment), 
% ((3,4),Entailment), 
% ((5,1),Contradiction), 
% ((5,\{6,7\}),Contradiction), 
% ((3,5),Neutral), 
% ((3,1),Neutral).

We designed a sequence of three experiments to evaluate the reasoning capabilities of our pipeline. First, we established a baseline by using the reformatted three-class dataset in our pipeline without generating any implicit premises. Second, we introduced single-step implicit premise generation 
%%via an LLM for entailment and contradiction classes providing the true label 
(as described in Section \ref{llmg}) to the pipeline. 
In the first and second experiments, we investigated different choices 
for the neuro-matching threshold $\tau_m$ and the neuro-contradiction threshold $\tau_c$.  
Third, using the optimal choices for $\tau_m$ and $\tau_c$ from the second experiment, 
we generated six-step implicit premise chains (as described in Section \ref{llmg})  and evaluated their performance using cumulative chains from 1 to 5 steps (i.e. first just step 1, second conjunction of steps 1 and 2, third conjunction of steps 1, 2 and 3, etc.) to analyze the impact of multi-step reasoning on classification performance. 
Note that the sixth step is always discarded because very often it is a redundant repetition of the claim. 
Table \ref{tab:exp_format} is an example of the data format (reformatted Microtext data and implicit premises) used across all three experiments.

\begin{table}[htbp]
    \centering
    \footnotesize % Reduces font size further
    \renewcommand{\arraystretch}{0.9} % Compresses vertical spacing between rows
    \begin{tabularx}{\columnwidth}{@{}lX@{}}
        \toprule
        \textbf{Reformatted} & \textbf{Premise:} Landfills also produce a lot of odor \\
        \textbf{Microtext} & \textbf{Claim:} Landfills are bad for handling our trash \\
       \textbf{data} &                      \textbf{Label:} Entailment \\
        \midrule
        \textbf{Experiment 1}     & \textbf{No Implicit Premise} \\
        \textbf{Experiment 2}     & \textbf{Single implicit premise:} Odor is a negative environmental impact. \\
        \textbf{Experiment 3}     & \textbf{Multi-step chain of implicit premises:} 
        (Step 1) Odor indicates poor waste decomposition; 
        (Step 2) Decomposition releases harmful gases; 
        (Step 3) These gases pollute the surrounding air; 
        (Step 4) Air pollution harms human health; 
        (Step 5) Health hazards show inadequate waste management; 
        (Step 6) Inadequate management makes landfills a bad solution. \\
        \bottomrule
    \end{tabularx}
    \caption{Example of the data format across all three experiments. }
    \label{tab:exp_format}
\end{table}

The reformatted and implicit data for each argumentative text is translated into AMR formulae, then into abstract formulae, and finally evaluated using automated reasoning, as explained for the pipeline in Section \ref{section:pipeline}. 

% To use this reformated and implicit data in our neuro-symbolic pipeline, the premise $\phi$ was the logical formula of the conjunction of the premise and implicit premise(s), and $\psi$ was the logical formula of the claim.}

%%%%%%%(Table \ref{tab:binary_example} shows the binary classification data transformed from Table \ref{tab:example}). 

\begin{figure}[h]
    \centering
    \begin{tikzpicture}
    \begin{axis}[
        width=0.9\textwidth,
        height=0.5\textwidth,
        xlabel={$\tau_m$},
        ylabel={Accuracy},
        legend style={at={(0.5,-0.2)}, anchor=north, legend columns=3},
        grid=major,
        xmin=0.5, xmax=0.8,
        ymin=0.33, ymax=0.52,
        xtick={0.5,0.55,0.6,0.65,0.7,0.75,0.8},
        ytick={0.33,0.36,0.39,0.42,0.45,0.48,0.51},
        tick label style={/pgf/number format/fixed},
    ]

    % Data for Experiment 2, tau_c = 80
    \addplot+[mark=*, blue, solid] coordinates {
        (0.5, 0.433333)
        (0.55, 0.464286)
        (0.6, 0.507143)
        (0.65, 0.476190)
        (0.7, 0.438095)
        (0.75, 0.402381)
        (0.8, 0.373810)
    };
    \addlegendentry{Exp2, $\tau_c=80$}

    % Data for Experiment 2, tau_c = 90
    \addplot+[mark=square*, red, solid] coordinates {
        (0.5, 0.433333)
        (0.55, 0.461905)
        (0.6, 0.511905)
        (0.65, 0.480952)
        (0.7, 0.433333)
        (0.75, 0.397619)
        (0.8, 0.366667)
    };
    \addlegendentry{Exp2, $\tau_c=90$}

    % Data for Experiment 2, tau_c = 100
    \addplot+[mark=triangle*, green, solid] coordinates {
        (0.5, 0.400000)
        (0.55, 0.430952)
        (0.6, 0.485714)
        (0.65, 0.450000)
        (0.7, 0.388095)
        (0.75, 0.364286)
        (0.8, 0.354762)
    };
    \addlegendentry{Exp2, $\tau_c=100$}

    % Data for Experiment 1, tau_c = 80
    \addplot+[mark=*, blue, dashed] coordinates {
        (0.5, 0.410072)
        (0.55, 0.426190)
        (0.6, 0.421429)
        (0.65, 0.371429)
        (0.7, 0.378571)
        (0.75, 0.354762)
        (0.8, 0.347619)
    };
    \addlegendentry{Exp1, $\tau_c=80$}

    % Data for Experiment 1, tau_c = 90
    \addplot+[mark=square*, red, dashed] coordinates {
        (0.5, 0.390476)
        (0.55, 0.411905)
        (0.6, 0.416667)
        (0.65, 0.366667)
        (0.7, 0.376190)
        (0.75, 0.352381)
        (0.8, 0.342857)
    };
    \addlegendentry{Exp1, $\tau_c=90$}

    % Data for Experiment 1, tau_c = 100
    \addplot+[mark=triangle*, green, dashed] coordinates {
        (0.5, 0.373810)
        (0.55, 0.390476)
        (0.6, 0.409524)
        (0.65, 0.361905)
        (0.7, 0.352381)
        (0.75, 0.338095)
        (0.8, 0.340476)
    };
    \addlegendentry{Exp1, $\tau_c=100$}

    \end{axis}
    \end{tikzpicture}
    \caption{Results for Experiment 1 (no implicit premises) and Experiment 2 (single implicit premise): 
    Accuracy across different $\tau_m$ and $\tau_c$ values (solid lines: Experiment 2; dashed lines: Experiment 1). As it is a three-class classification task, random classification would tend to have an accuracy of 0.333.}
    \label{fig:merged_accuracy}
\end{figure}
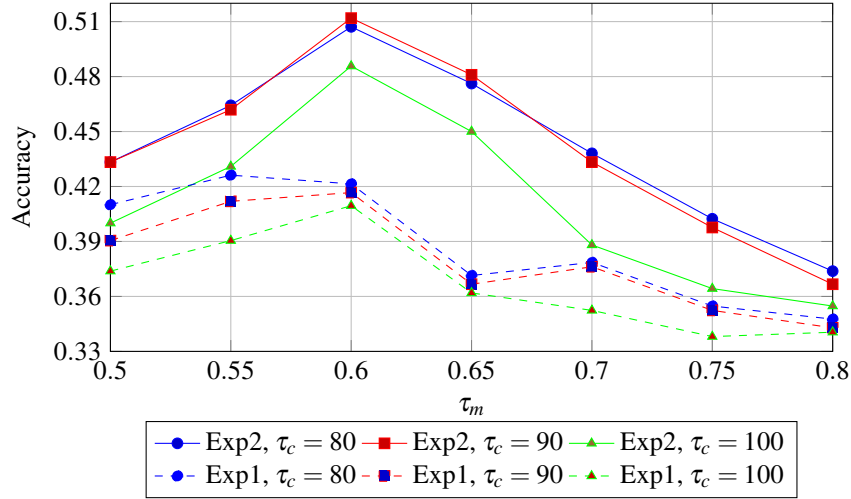
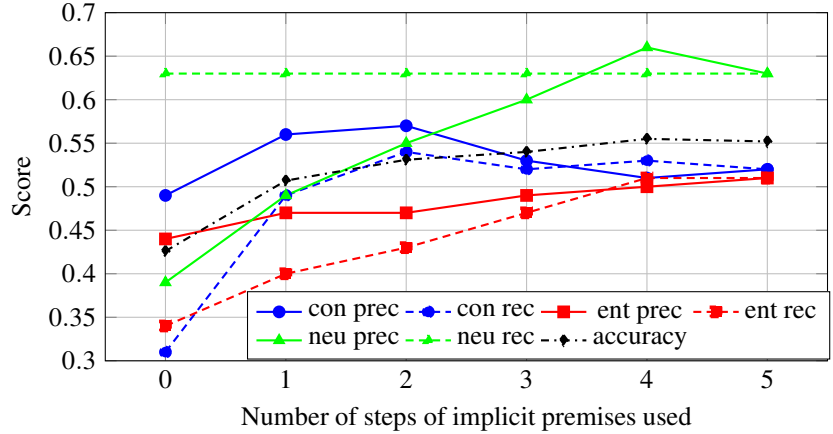
\begin{figure}[h]
    \centering
    \begin{tikzpicture}
    \begin{axis}[
        width=0.9\textwidth,
        height=0.5\textwidth,
        xlabel={Number of steps of implicit premises used},
        ylabel={Score},
        legend style={at={(0.6,0.2)}, anchor=north, legend columns=4},
        grid=major,
        xmin=-0.5, xmax=5.5,
        ymin=0.3, ymax=0.7,
        xtick={0,1,2,3,4,5},
        ytick={0.3,0.35,0.4,0.45,0.5,0.55,0.6,0.65,0.7},
        ymajorgrids=true,
    ]

    % con precision
    \addplot[blue, mark=*, thick, solid] coordinates {
        (0,0.49) (1,0.56) (2,0.57) (3,0.53) (4,0.51) (5,0.52)
    };
    \addlegendentry{con prec}

    % con recall
    \addplot[blue, mark=*, thick, densely dashed] coordinates {
        (0,0.31) (1,0.49) (2,0.54) (3,0.52) (4,0.53) (5,0.52)
    };
    \addlegendentry{con rec}

    % ent precision
    \addplot[red, mark=square*, thick, solid] coordinates {
        (0,0.44) (1,0.47) (2,0.47) (3,0.49) (4,0.50) (5,0.51)
    };
    \addlegendentry{ent prec}

    % ent recall
    \addplot[red, mark=square*, thick, densely dashed] coordinates {
        (0,0.34) (1,0.40) (2,0.43) (3,0.47) (4,0.51) (5,0.51)
    };
    \addlegendentry{ent rec}

    % neu precision
    \addplot[green, mark=triangle*, thick, solid] coordinates {
        (0,0.39) (1,0.49) (2,0.55) (3,0.60) (4,0.66) (5,0.63)
    };
    \addlegendentry{neu prec}

    % neu recall
    \addplot[green, mark=triangle*, thick, densely dashed] coordinates {
        (0,0.63) (1,0.63) (2,0.63) (3,0.63) (4,0.63) (5,0.63)
    };
    \addlegendentry{neu rec}

    % accuracy
    \addplot[black, mark=diamond*, thick, dashdotted] coordinates {
        (0,0.426) (1,0.507) (2,0.531) (3,0.540) (4,0.555) (5,0.552)
    };
    \addlegendentry{accuracy}

    \end{axis}
    \end{tikzpicture}
    \caption{Experiment 3 (using 1 to 5 steps of implicit premises): Precision (prec) and recall (rec) scores for all three classes and overall accuracy across different steps ($\tau_m$=0.6, $\tau_c$=80).}
    \label{fig:step_analysis}
\end{figure}

\paragraph{Results}
For all three experiments, we used the same subset of the reformatted data (as described in the methods), comprising 140 randomly selected items per class (420 items total). In the first two experiments, we evaluated performance across a range of neuro-matching thresholds \(\tau_m\) (from 0.5 to 0.8) and three neuro-contradiction thresholds \(\tau_c\) (viz. 80, 90, and 100).  

As shown in Figure~\ref{fig:merged_accuracy}, a comparative analysis of model performance reveals distinct optimal thresholds for each experiment. Experiment 1 achieved peak accuracy ($\sim$0.426) at \(\tau_m = 0.55\) and \(\tau_c = 80\), while Experiment 2 attained its highest accuracy ($\sim$0.512) at \(\tau_m = 0.6\) and \(\tau_c = 90\). In both cases, accuracy declined steadily as \(\tau_m\) increased beyond these values. The parameter \(\tau_c\) exhibited a modest influence on performance, with lower values generally yielding marginally better results. Class-level F1-scores revealed important patterns: in Experiment 1, the \textit{con} and \textit{ent} classes performed best at \(\tau_m = 0.55\), while in Experiment 2, these classes showed good F1 scores within the range \(\tau_m = 0.5\)–0.6. The \textit{neu} class demonstrated stronger F1 performance at higher thresholds, particularly beyond \(\tau_m = 0.6\). Notably, lower \(\tau_c\) values resulted in better \textit{con} performance without significantly degrading the \textit{ent} results and \textit{neu} results. 

Based on the above findings, we selected \(\tau_m = 0.6\) and \(\tau_c = 80\) (rather than \(\tau_c = 90\)) for Experiment 3 to prioritize better performance for both \textit{con} and \textit{ent} classes while maintaining reasonable overall accuracy.
%%%%%%This configuration was used to evaluate iterative refinement performance under stable threshold conditions. 
As shown in Figure~\ref{fig:step_analysis}, the use of an increasing number of reasoning steps as implicit premises leads to consistent improvements in both class-specific metrics and overall accuracy under fixed thresholds ($\tau_m = 0.6$, $\tau_c = 80$). The \textit{con} class demonstrates a particularly notable improvement: precision increases from 0.49 to 0.52, while recall nearly doubles from 0.31 to 0.52, indicating substantially improved detection capability. Similarly, the \textit{ent} class exhibits steady gains, with precision and recall rising from approximately 0.44 and 0.34, respectively, to 0.51. The \textit{neu} class shows a different pattern: while recall remains consistently high at 0.63 across all step counts because we do not generate implicit-premise steps for the neutral class, precision improves markedly from 0.39 to 0.63, reflecting enhanced prediction confidence for this majority class. Overall accuracy increases from 0.426 to 0.555, reaching its peak at 4 steps before a slight decrease at 5 steps, suggesting optimal performance is achieved with four steps of implicit premise integration. 
% The convergence of precision and recall values across all classes by 5 steps indicates \myred{development toward a more balanced classifier}, with 4 steps providing the best trade-off between overall accuracy and metric stability. These results underscore the value of multi-step reasoning for enhancing model performance, particularly for minority classes.
By step 5, precision and recall converge across all classes, indicating a more balanced classifier. However, using 4 steps provides the optimal trade-off between overall accuracy and metric stability. These results demonstrate that multi-step reasoning improves model performance, especially for minority classes.

\section{Discussion}
\label{section:discussion}
%%%%%%%%%%%%%%%%%%%%

This paper presents the first framework to systematically generate implicit premises that logically justify a known semantic relation (e.g., entailment or contradiction) between an explicit premise and a claim. Given a premise-claim pair and its relation label, our method identifies the implicit premise that makes the logical connection explicit.
We evaluated our pipeline on a dataset with substantial implicit reasoning. The approach achieved strong, balanced performance across all three classes in classification tasks, with the additional benefit of providing logically reconstructed arguments.
Whilst the results are promising, there is a need to further improve the pipeline's performance, perhaps by focusing on better ways to prompt an LLM for appropriate implicit premises. 

In future work, we will investigate using this pipeline for end-to-end logical argument mining by directly processing plain text to identify relations through neuro-symbolic translation, thereby enabling the systematic generation of logical argument graphs from unstructured text.

\bibliographystyle{vancouver}
\bibliography{comma}

%%%%%%%%%%%%%%%%%%%%%%%%%%%%%%%%%%%%%%%%%%%%%%
%%%%%%%%%%%%%%%%%%%%%%%%%%%%%%%%%%%%%%%%%%%%%%
\clearpage

\end{document}